\documentclass[letterpaper, 10 pt, conference]{ieeeconf}  % Comment this line out if you need a4paper

\IEEEoverridecommandlockouts                              % This command is only needed if 
\usepackage{times}

\usepackage[bookmarks=true]{hyperref}
\usepackage{multicol}
\usepackage{booktabs}
\usepackage[weather]{ifsym}
\usepackage{siunitx}
\usepackage{tabularx}

\usepackage[dvipsnames]{xcolor}
\usepackage{moreverb,url}
\usepackage{amsmath,amsfonts,amssymb}
\usepackage{tikz,tkz-euclide}
\usepackage{balance}
\usetikzlibrary{decorations,calligraphy}
\usetikzlibrary{calc}
\tikzset{>=latex}
\usepackage{multicol}
\usepackage{multirow}
\usepackage{tabularx}
\usepackage{booktabs}
\usepackage{siunitx}
\newcolumntype{Y}{>{\centering\arraybackslash}X}

\usepackage[nolist,nohyperlinks]{acronym}

\def\m{m}

\renewcommand\time[1]{{t_{#1}}}

\newcommand\imutime[1]{{\tau_{#1}}}

\newcommand\trans[2]{{\mathbf{T}_{#1}^{#2}}}
\newcommand\transplanar[2]{{\bar{\mathbf{T}}_{#1}^{#2}}}

\newcommand\rot[2]{{\mathbf{R}_{#1}^{#2}}}
\newcommand\rotplanar[2]{{\bar{\mathbf{R}}_{#1}^{#2}}}

\newcommand\pos[2]{{\mathbf{p}_{#1}^{#2}}}
\newcommand\posplanar[2]{{\bar{\mathbf{p}}_{#1}^{#2}}}

\newcommand\vel[2]{{\mathbf{v}_{#1}^{#2}}}
\newcommand\velz[2]{{{v}_{z#1}^{#2}}}
\newcommand\velplanar[2]{{\bar{\mathbf{v}}_{#1}^{#2}}}

\def\world{{W}}

\newcommand\cartcoords[1]{{\mathbf{X}_{#1}}}
\newcommand\cartcoord[1]{{\mathbf{x}_{#1}}}
\def\crosscor{{f}}
\def\undistfunc{{u}}
\def\dopplerfunc{{d}}
\def\Exp{\text{Exp}}
\def\Expplanar{\text{Exp}}

\newcommand\gyr[1]{{\boldsymbol{\omega}_{#1}}}
\newcommand\gyryaw[1]{{\bar{\omega}_{#1}}}
\newcommand\gyrplanarset[1]{{\bar{\boldsymbol{\omega}}_{#1}}}

\newcommand\argmax[1]{{\underset{#1}{\text{argmax}}}}

\newcommand\intensity[1]{{\psi_{#1}}}

\newcommand\image[1]{{\mathcal{I}^{#1}}}

\def\bilinear{\beta}

\def\localmap{\mathcal{M}}

\begin{acronym}[AAAAAAAAA]
    \acro{dof}[DoF]{Degree-of-Freedom}
    \acro{fov}[FoV]{Field-of-View}
    \acro{icp}[ICP]{Iterative Closest Point}
    \acro{rmse}[RMSE]{Root Mean Square Error}
    \acro{slam}[SLAM]{Simultaneous Localization And Mapping}
\end{acronym}

\title{\LARGE \bf
3DRO: Lidar-level SE(3) Direct Radar Odometry \\Using a 2D Imaging Radar and a Gyroscope 
}

\author{Cedric Le Gentil$^{1,2}$, Daniil Lisus$^{1}$, Timothy D. Barfoot$^{1}$% <-this % stops a space
\thanks{$^{1}$~Robotics Institute, University of Toronto.
        {Corresponding author: \tt\small cedric.legentil@robotics.utias.utoronto.ca}}%
\thanks{$^{2}$~Mobile Robotics Lab, ETH Zurich.}
}

\begin{document}

\maketitle
\thispagestyle{empty}
\pagestyle{empty}

%%%%%%%%%%%%%%%%%%%%%%%%%%%%%%%%%%%%%%%%%%%%%%%%%%%%%%%%%%%%%%%%%%%%%%%%%%%%%%%%
\begin{abstract}

Recently, the robotics community has regained interest in radar-based perception and state estimation.
A 2D imaging radar provides dense 360$^\circ$ information about the environment.
Despite the radar antenna's cone of emission and reception, the collected data is generally assumed to be limited to the plane orthogonal to the radar’s spinning axis.
Accordingly, most methods based on 2D imaging radars only perform SE(2) state estimation.
This paper presents 3DRO, an extension of the SE(2) Direct Radar Odometry (DRO) framework to perform state estimation in SE(3).
While still assuming planarity of the data through DRO’s 2D velocity estimates, it integrates 3D gyroscope measurements over SO(3) to estimate SE(3) ego motion.
While simple, this approach provides lidar-level odometry accuracy as demonstrated using 643km of data from the Boreas-RT dataset.

\end{abstract}

\section{Introduction}

Radars play a significant role in the automotive, aerospace, and maritime industries.
A crucial advantage of radar sensing is its robustness to weather conditions and ability to provide Doppler-based velocity measurements.
The robotics research community is regaining interest in such a modality, as illustrated by a recent survey~\cite{harlow2024newwave}.
Two main types of radar sensors are commonly used in robotics: 2D imaging and 3D/4D radars.
The former generally consists of a single rotating antenna, providing a 360\textdegree{} \ac{fov} in the horizontal plane, while the latter provides 3D point clouds with radial velocities over a limited azimuth \ac{fov} but over multiple elevations, typically using phased-array antennas.
A common critique against 2D imaging radars is their lack of elevation information, thus limiting their use to SE(2) state estimation problems.
In this work, we investigate the use of a 2D imaging radar, in combination with a gyroscope, to perform SE(3) odometry for ground robots.

\begin{figure}
    \def\legenddist{0.1cm}
    \def\dataheight{3.0cm}
    \def\vdist{3.8cm}
    \centering
    \begin{tikzpicture}
        \tikzstyle{legend} = [align = center, inner sep=0, outer sep=0, node distance = 0em, execute at begin node=\setlength{\baselineskip}{8pt}\small] 
        \node[inner sep = 0, outer sep = 0] (boreas) {\includegraphics[clip, trim=1.5cm 0cm 0cm 0cm, width = \columnwidth]{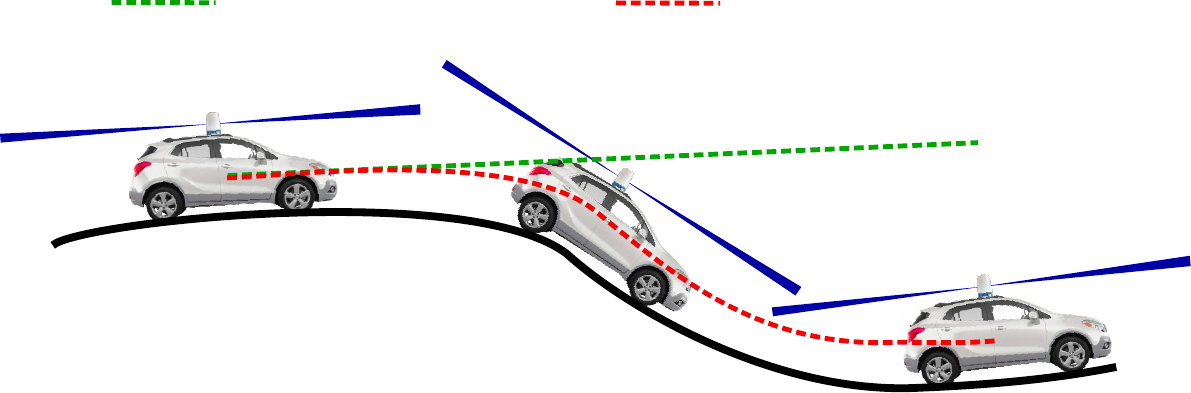}};
        \node[] at (-2.2,1.5) {\small SE(2) odometry};
        \node[] at (1.75,1.5) {\small SE(3) odometry};
        \node[legend, below=\legenddist of boreas] {(a) Illustration of 2D radar sensing in a 3D world};
        \node[inner sep=0, outer sep=0, below=\vdist of boreas.west, anchor=west] (polar){\includegraphics[clip, height = \dataheight]{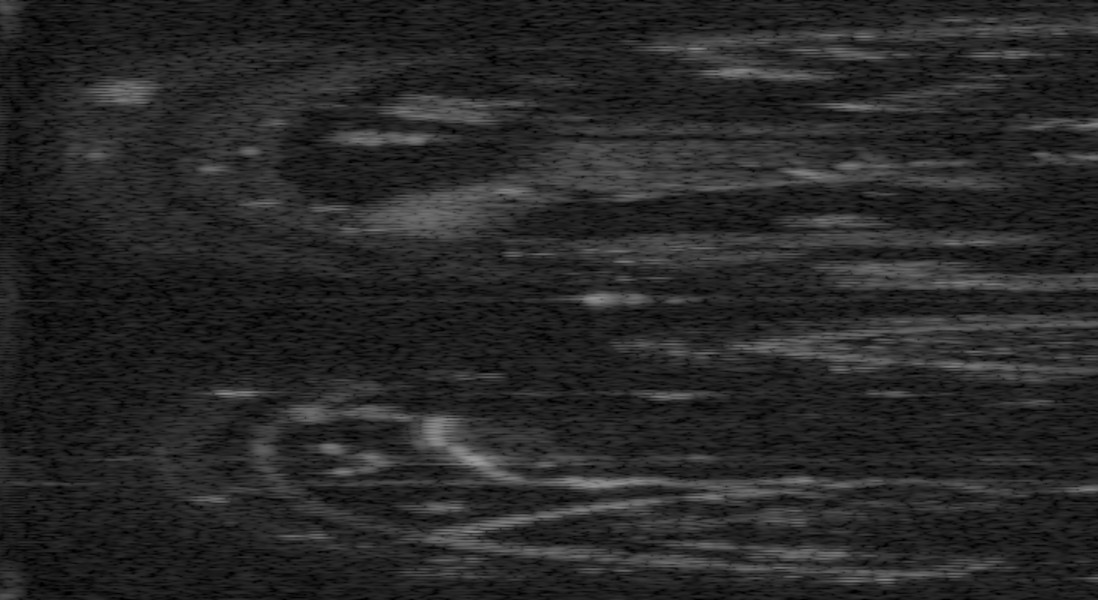}};
        \node[inner sep=0, outer sep=0, below=\vdist of boreas.east, anchor=east] (cart){\includegraphics[clip, height = \dataheight]{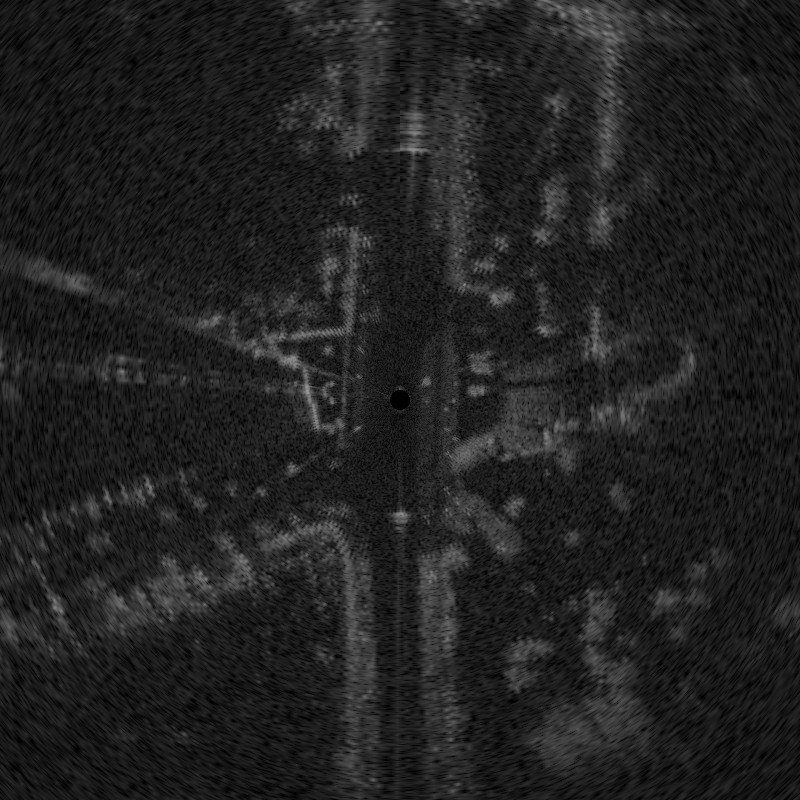}};
        \node[legend, below=\legenddist of polar] {(b) Raw radar polar image};
        \node[legend, below=\legenddist of cart] {(c) Cartesian image};
    \end{tikzpicture}
    %\vspace{-0.3cm}
    \caption{Imaging radars collect data by sweeping radio waves around and analyzing the returned signal. Due to the narrow cone of emission, this type of device is generally considered as a 2D sensor. (b) and (c) present typical data obtained with a 2D imaging radar in raw/polar and Cartesian forms, respectively. In this work, we investigate the problem of SE(3) state estimation using a 2D imaging radar in combination with a 3-DoF gyroscope. Our experiments suggest that such a sensor suite can provide lidar-level odometry accuracy.}
    \label{fig:teaser}
\end{figure}

As shown in Fig.~\ref{fig:teaser}, a scan collected by a 2D imaging radar consists of a polar image, where each column corresponds to a range bin and each row corresponds to an azimuth bin.
The value of each pixel corresponds to the intensity of the radar return.
Overall, the intensity image provides the full \emph{waveform} of the signal.
Commonly, radar state estimation extracts features from the raw scan and forms point clouds before performing association-based or \ac{icp}-like scan matching.
This strategy lowers the computational cost of state estimation, but it also discards a significant amount of information contained in the raw data.
Another line of work performs scan registration using the full radar intensity image to leverage the rich information contained in the radar returns.
These methods are known as \emph{direct} methods.
DRO~\cite{legentil2025dro} recently demonstrated state-of-the-art performance for SE(2) radar odometry combining gyroscope yaw measurement integration and direct scan registration with a constant-velocity motion model.
This paper presents 3D Direct Radar Odometry (3DRO), an extension of DRO~\cite{legentil2025dro} to SE(3) state estimation by integrating gyroscope measurements over SO(3).

Part of the challenge to perform SE(3) odometry using DRO-generated 2D velocities and readings from a 3-\ac{dof} gyroscope is the extrinsic calibration between the sensors and the moving platform.
The velocity estimates in 2D do not capture the full 3D motion of the platform if the radar sensing plane is not parallel to the plane of motion.
In simpler words, if the radar is tilted with respect to the ground, the sensor velocity on the vertical axis will be non-zero and cannot be observed using a 2D radar measurement model.
To address this issue, we propose a simple extrinsic calibration method to effectively approximate the vertical velocity of the sensor and enable lidar-level SE(3) odometry.
We evaluate our method using the Boreas-RT dataset~\cite{lisus2026boreasrt}, which includes around \SI{650}{\kilo\meter} of data collected with an automotive platform in various environments.

\section{Related work}

Radar sensors used for robotic perception broadly fall into two categories: phased-array systems for 3D/4D data collection, and mechanically spinning antennas for \SI{360}{\degree} 2D sensing.
This section provides a brief literature review of state estimation methods based on the latter.
For a broader overview of 3D/4D radar perception, we refer the reader to~\cite{harlow2024newwave}, \cite{nader2024survey}, and~\cite{venon2022millimeter}.

Most radar-based state estimation methods operate on point clouds or features extracted from raw polar or Cartesian data.
Early approaches, such as~\cite{callmer2011radarslamusingvisualfeatures}, apply standard computer-vision keypoints and descriptors~\cite{lowe2004sift} directly to Cartesian radar images.
Later methods extract points per azimuth from the polar image using detectors such as CFAR~\cite{rohling1983radarcfar}, K-strongest~\cite{adolfsson2021CFEAR}, or CFEAR~\cite{adolfsson2023cfear}, and the impact of these choices on odometry performance was systematically evaluated in~\cite{preston2025finer}.
A key characteristic of spinning radar data is that scans are not collected instantaneously: with the antenna sweeping at approximately \SI{4}{\hertz}, platform motion during acquisition causes motion distortion.
This distortion has two components: the displacement of the sensor between consecutive azimuths, and Doppler-induced range shifts arising from the relative velocity between the sensor and the environment.
One remedy is a one-off undistortion step based on prior state estimates, as in~\cite{adolfsson2023cfear}.
Alternatively, continuous-time state estimation jointly handles undistortion and ego-motion estimation, as demonstrated by~\cite{burnett2021dowe}, \cite{are_we_ready_for}, and~\cite{burnett2025continuous}.
Robustness can further be improved through learned feature representations~\cite{hero_paper} or learned weighting schemes~\cite{Lisus2025Pointing}.

Beyond odometry, radar data is also used in full-batch trajectory optimization with loop-closure constraints to mitigate accumulated drift.
In this setting, the trajectory is estimated by optimizing a pose graph where each edge encodes a constraint on the relative pose between two timestamps.
Radar \ac{slam} systems such as~\cite{holder2019realtime}, \cite{hong2022radarslam}, and~\cite{adolfsson2023tbv} adopt this approach, relying on loop-closure detectors originally designed for lidar data (\cite{himstedt2014largescale}, \cite{he2016m2dp}, and~\cite{kim2022scancontextpp}, respectively).

A parallel line of radar works uses \emph{direct} methods for scan registration.
They operate on the full radar intensity image without extracting features.
The most common approach evaluates cross-correlation scores between the current scan and a reference over a discrete set of candidate state updates, before selecting the highest-scoring hypothesis.
Works such as~\cite{Checchin2009}, \cite{masking_by_moving}, and~\cite{park2020pharao} follow this principle, though without explicitly modelling motion or Doppler distortion.
The authors of~\cite{lisus2025doppler} showed that Doppler distortion can itself serve as a reliable odometry source in feature-deprived environments, using a radar with triangular modulation paired with a yaw-gyroscope.
DRO~\cite{legentil2025dro} is the first continuous direct method to rigorously account for both motion and Doppler distortions, maximizing a continuous cross-correlation score between the current scan and a local map updated on-the-fly.
It has recently been integrated in a \ac{slam} framework named Dr-PoGO~\cite{legentil2026drpogo} that uses RaPlace~\cite{jang2023raplace} for loop-closure detection, and a variant of DRO for loop registration.
3DRO builds on DRO by integrating full 3-\ac{dof} gyroscope measurements to extend direct radar odometry from SE(2) to SE(3), enabling lidar-level performance based on a 2D imaging radar.

\section{Background - SE(2) direct radar odometry}

For completeness, this section provides a high-level overview of DRO~\cite{legentil2025dro} for SE(2) odometry.
Let us consider a 2D imaging radar that provides scans as polar images $\image{i}$ ($i = 1, \cdots, N$), rigidly mounted on a platform with a gyroscope that provides angular velocity measurements $\gyr{j}$ ($j = 1, \cdots, M$).
Without loss of generality, we assume that the gyroscope measurements are transformed to the radar frame according to the extrinsic calibration between the two sensors.
We denote the yaw component of the gyroscope measurements as $\gyryaw{j}$.
DRO estimates the SE(2) pose of the sensor by performing scan-to-local-map registration.
The rotational component of the motion is obtained by integrating the gyroscope's yaw measurements $\gyryaw{j}$, while the translational component is estimated under the assumption of constant linear body  $\velplanar{i}{}$ by registering the current scan $\image{i}$ to a local map $\localmap_{i-1}$.
The SE(2) pose $\transplanar{\world}{\time{}}$ of the sensor at time $\time{} \in [\time{i-1},\time{i}]$ is obtained as
\begin{equation}
    \transplanar{\world}{\time{}} = \transplanar{\world}{\time{i-1}} \transplanar{\time{i-1}}{\time{}},
\end{equation}
where $\transplanar{\time{i-1}}{\time{}}=\begin{bmatrix}\rotplanar{\time{i-1}}{\time{}} & \posplanar{\time{i-1}}{\time{}} \\ 0 & 1\end{bmatrix}$ is the pose change relative to the previous time step $\time{i-1}$.
This transformation is obtained as
\begin{align}
    \rotplanar{\time{i-1}}{\time{}} &= \Expplanar\left(\int_{\time{i-1}}^{t} \gyryaw{s} ds\right),
    \label{eq:rotmotionmodel}
    \\
    \posplanar{\time{i-1}}{\time{}} &= \int_{\time{i-1}}^{t} \rotplanar{\time{i-1}}{s} \velplanar{i}{} ds,
    \label{eq:motionmodel}
\end{align}
with $\Expplanar(\cdot)$ the exponential map from $\mathfrak{so}(2)$ to $\mathrm{SO}(2)$.
Note that the integration of the gyroscope measurements (right-hand side of~\eqref{eq:rotmotionmodel}) can be performed using standard Riemann integration in $\mathfrak{so}(2)$ before using the exponential map, as rotations are commutative in 2D, which is not the case in 3D.
The linear body velocity $\velplanar{i}{}$ is the state variable to be estimated.
The continuous nature of~\eqref{eq:motionmodel} allows for the undistortion of the radar scans, accounting for the motion of the platform during data acquisition by applying a different transformation to each row/azimuth of the scan.

The function $\dopplerfunc(\image{i}, \velplanar{i}{})$ corrects the Doppler distortion of the raw radar scan $\image{i}$, by shifting the range bins of the scan according to the body velocity $\velplanar{i}{}$\footnote{For a detailed description of the Doppler correction, please refer to~\cite{legentil2025dro}}.
The function $\undistfunc(\image{i}, \gyrplanarset{i}, \velplanar{i}{})$ performs the scan undistortion and conversion from polar to Cartesian coordinates.
The vector $\gyrplanarset{i}$ is the set of gyroscope yaw measurements which occurred during the acquisition of the scan $\image{i}$.
Given the current velocity estimate, the raw radar scan $\image{i}$ is first Doppler-corrected and then undistorted to obtain Cartesian coordinates $\cartcoords{i}$ in the local map frame at time $\time{i-1}$:
\begin{equation}
     \cartcoords{i} = \undistfunc\left(\dopplerfunc\left(\image{i}, \velplanar{i}{}\right), \gyrplanarset{i}, \velplanar{i}{}\right).
\end{equation}
Each Cartesian point in $\cartcoords{i}$ has a corresponding intensity $\intensity{j}$, inherited from the original data.
The estimation of the velocity $\velplanar{i}{}$ is performed by maximizing a continuous cross-correlation score $\crosscor$ between the intensity values $\intensity{j}$ of the Cartesian coordinates $\cartcoords{i}$ and the corresponding location in the local map $\localmap_{i-1}$.
The latter consists of a Cartesian image generated by aggregating the previous scans, as detailed later in this section.
Bilinear interpolation $\bilinear(\cdot)$ is used to obtain the intensity values of the local map at the Cartesian coordinates $\cartcoords{i}$.
The optimization problem is thus defined as
\begin{equation}
    \velplanar{i}{*} = \argmax{\velplanar{i}{}} \sum_{\cartcoord{j} \in \cartcoords{i}} \intensity{j} \bilinear\left(\localmap_{i-1}, \cartcoord{j}\right).
    \label{eq:droopt}
\end{equation}
DRO~\cite{legentil2025dro} also includes a special Doppler-based velocity constraint to improve robustness in feature-deprived environments.
We refer the reader to~\cite{legentil2025dro} for more details on the full optimization problem.

As mentioned earlier, the local map consists of a Cartesian image $\localmap_{i-1}$ that combines information from the previous scans.
Concretely, it is a regular grid of low-pass filters that are updated with the undistorted version of the current scan $\image{i}$ after solving~\eqref{eq:droopt}.
The local map is moved at each step to keep it centred on the current sensor position.
To summarize, DRO's optimization provides 2D velocity estimates $\velplanar{i}{}$ in the plane of the radar scan by performing scan registration using a motion model based on velocity-gyroscope integration.

\section{Method}

\subsection{Vertical velocity approximation}

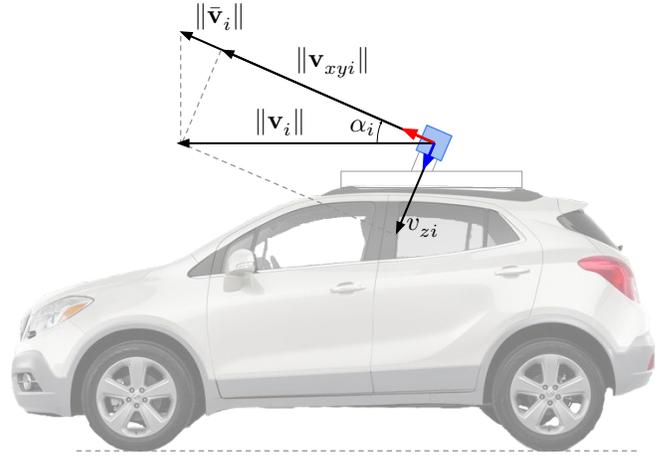
\begin{figure}
    \centering
    \begin{tikzpicture}
    \node{\includegraphics[clip,width=0.99\columnwidth]{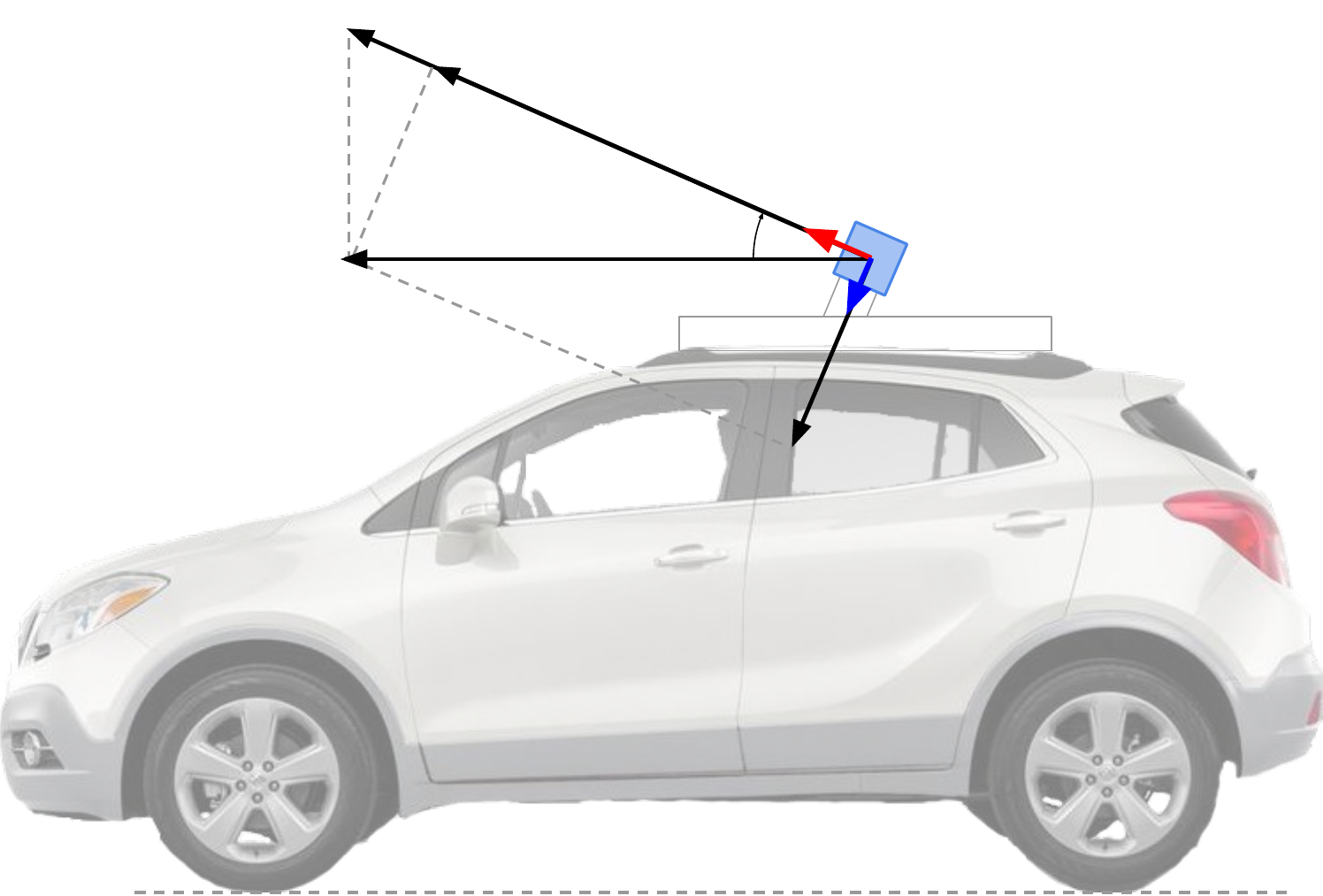}};
    \node at (-1.5,2.9) {$\Vert\velplanar{i}{}\Vert$};
    \node at (-0.0,2.3) {$\Vert\vel{xyi}{}\Vert$};
    \node at (1.2,0.1) {$\velz{i}{}$};
    \node at (-0.7,1.5) {$\Vert\vel{i}{}\Vert$};
    \node at (0.4,1.42) {$\alpha_i$};
    \end{tikzpicture}
    \caption{Illustration of the misalignment between the radar sensing plane and the plane of motion. Note that the angle $\alpha_i$ is amplified for clarity. In reality, this angle is close to zero.}
    \label{fig:boreas_radar}
\end{figure}

As mentioned in the introduction, the main challenge to perform SE(3) odometry using DRO-generated 2D velocities and readings from a 3-\ac{dof} gyroscope is the incomplete nature of the velocity estimates if the radar sensing plane is not parallel to the plane of motion.
Fig.~\ref{fig:boreas_radar} provides an exaggerated illustration of the phenomenon with an angle misalignment $\alpha_i$ around the pitch axis.
The velocity estimated with a planar motion assumption $\velplanar{i}{}$ is not colinear with the radar velocity $\vel{i}{}=\begin{bmatrix}\vel{xyi}{\top}&\velz{i}{}\end{bmatrix}^\top$.
However, for the particular example illustrated in Fig.~\ref{fig:boreas_radar}, we can easily link them through simple trigonometry with $\Vert\vel{xyi}{}\Vert~=~\cos(\alpha_i)^2\Vert\velplanar{i}{}\Vert$ and $\velz{i}{}=\sin(\alpha_i)\cos(\alpha_i)\Vert\velplanar{i}{}\Vert$.

It is important to note that, due to the non-rigid nature of the platform (suspension, tyre deformation, etc.), the angle $\alpha_i$ is not constant through time.
The sensor inclination with respect to the plane of motion is impacted by various factors such as the roll and pitch of the vehicle's body, the terrain and acceleration profiles, etc.
Nonetheless, in this work, we approximate it as constant.
Thus, we simplify the expression of $\velz{i}{}$ as
\begin{equation}
    \velz{i}{} = \kappa \Vert\velplanar{i}{}\Vert,
    \label{eq:velz}
\end{equation}
with $\kappa$ a calibration constant.

The calibration of $\kappa$ is performed by leveraging the ground-truth body velocity $\vel{i}{\rm{GT}}$ of the radar in 3D.
The value of $\kappa$ is obtained as
\begin{equation}
    \kappa = \frac{1}{N} \sum_{i=1}^{N} \frac{\velz{i}{\rm{GT}}}{\Vert{\velplanar{i}{}}\Vert}.
    \label{eq:alpha}
\end{equation}
Due to the norm in the denominator, this calibration assumes that the vehicle is always moving in the same direction.
Adding a sign that accounts for the direction of motion in~\eqref{eq:velz} and ~\eqref{eq:alpha} would make the method more general.
In our implementation, we also avoid numerical issues by only including time steps where the norm of the planar velocity is above \SI{3}{\meter\per\second}.
Calibrating using only one sequence of the Boreas-RT dataset, we obtained the value of $\kappa=5.9\times 10^{-3}$, which corresponds to $\alpha_i\approx0.34^\circ$.
In that context, the $xy$ components of the radar velocities can safely be approximated as $\vel{xyi}{}\approx\velplanar{i}{}$.
Our experiments demonstrate that these simple approximations are sufficient to achieve lidar-level SE(3) odometry performance in various environments.

\subsection{SE(3) odometry}

The SE(3) pose $\trans{\world}{\imutime{j}}$ of the sensor at the gyroscope timestamp $\imutime{j}$ is obtained by integrating the velocity estimates $\vel{i}{}$ and the gyroscope measurements $\gyr{j}$ (with $\imutime{j} \in [\time{i-1}, \time{i}]$) as
\begin{equation}
    \trans{\world}{\imutime{j}} = \trans{\world}{\imutime{j-1}} \begin{bmatrix}\rot{\imutime{j-1}}{\imutime{j}} & \pos{\imutime{j-1}}{\imutime{j}} \\ 0 & 1\end{bmatrix}
    \label{eq:se3int}
\end{equation}    
with
\begin{align}
    \rot{\imutime{j-1}}{\imutime{j}} &= \Exp\left(\frac{\gyr{j} + \gyr{j-1}}{2} (\imutime{j} - \imutime{j-1})\right), \\
    \pos{\imutime{j-1}}{\imutime{j}} &= \vel{i}{} (\imutime{j} - \imutime{j-1}),
\end{align}
and $\Exp(\cdot)$ the exponential map from $\mathfrak{so}(3)$ to $\mathrm{SO}(3)$.

These derivations assume that the gyroscope measurements are unbiased.
In practice, as done in~\cite{legentil2025dro}, a simple bias estimation strategy consists of averaging the gyroscope measurements when the platform is stationary (i.e., when the velocity estimates are close to zero).
Then the estimated biases are subtracted from the gyroscope measurements before performing the integration.

\section{Experiments}

We evaluate 3DRO on the Boreas-RT dataset~\cite{lisus2026boreasrt}, which contains around \SI{650}{\kilo\meter} of data collected with an automotive platform in Southern Ontario, Canada.
The dataset consists of 60 sequences repeatedly collected over 9 different routes.
The sensor suite includes a Navtech RAS6 imaging radar, a 6-\ac{dof} Silicon Sensing DMU41 IMU, and a Velodyne Alpha Prime lidar.
The ground truth is obtained using a high-precision Applanix RTK-GNSS-INS system.
We compute the KITTI odometry metrics with the dataset-provided evaluation scripts for both SE(3) and SE(2) odometry.
The calibration of $\kappa$ is obtained using a single sequence (\texttt{2024-12-03-12-54}) from the \texttt{suburbs} route type.
Our implementation is available at \url{https://github.com/utiasASRL/dro}.

\subsection{SE(3) odometry results}

\begin{figure}
    \centering
    \includegraphics[width=0.99\columnwidth]{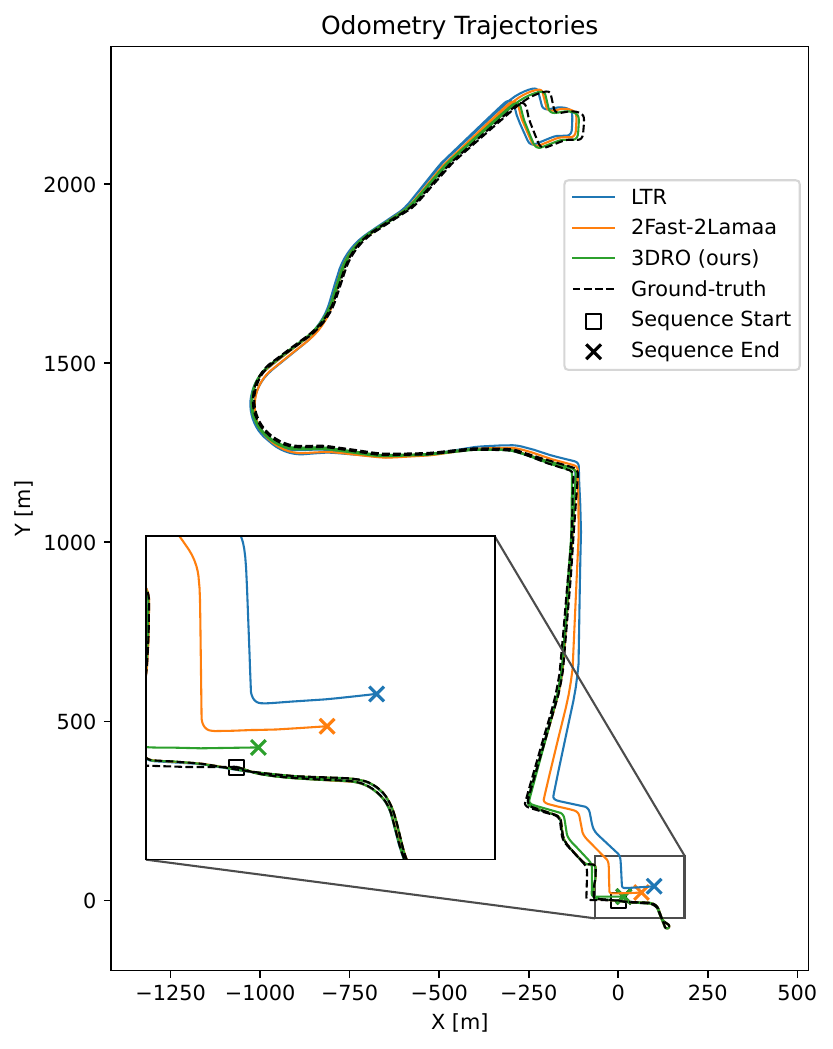}
    \caption{Example of odometry trajectories obtained with 3DRO and the various baselines of our SE(3) performance analysis.}
    \label{fig:odom_traj}
\end{figure}

\begin{table*}
    \centering
    \caption{Per-route average SE(3) odometry accuracy on the Boreas-RT dataset (\textbf{best} and \underline{second best}).}
    \setlength{\tabcolsep}{2pt}
\label{tab:odomse3}
    \begin{tabularx}{\linewidth}{lYYYYY}
\toprule
\textbf{Seq. type} & \textbf{LTR w/ gyro} \cite{are_we_ready_for} & \textbf{2Fast-2Lamaa} \cite{legentil20262fast} & \textbf{DRO} \cite{legentil2025dro} & \textbf{3DRO} No $\velz{}{}$& \textbf{3DRO} (ours)\\
%\multirow{2}{*}{\textbf{Seq. type}} & \textbf{LTR/ gyro} & \textbf{2Fast-} & \textbf{3DRO} & \textbf{3DRO}\\
% & [3] & \textbf{2lamaa} [4] & No $\velz{}{}$ & (ours)\\
\midrule

\texttt{suburbs}    & \underline{0.29}/\underline{0.09} & \textbf{0.22}/\textbf{0.08} & 1.38/0.58 & 0.55/\underline{0.09} & 0.31/\underline{0.09} \\
\texttt{industrial} & \underline{0.25}/\underline{0.09} & \textbf{0.20}/\textbf{0.07} & 1.41/0.63 & 0.66/0.13 & 0.49/0.13 \\
\texttt{urban}      & \underline{0.49}/\underline{0.20} & \textbf{0.38}/\textbf{0.10} & 1.54/0.66 & 1.01/0.32 & 0.92/0.32 \\
\texttt{forest}     &             0.70/\underline{0.11} & \underline{0.47}/0.18 & 3.04/1.22 & 0.51/\textbf{0.08} & \textbf{0.34}/\textbf{0.08} \\
\texttt{farm}       & \underline{0.44}/\underline{0.12} & \textbf{0.36}/\textbf{0.10} & 1.09/0.48 & 0.62/0.15 & 0.54/0.15\\
\texttt{regional}   & \underline{0.35}/\underline{0.09} & \textbf{0.20}/\textbf{0.08} & 1.19/0.49 & 0.63/\textbf{0.08} & 0.39/\textbf{0.08} \\
\texttt{tunnel}     & 2.06/\underline{0.10} & \textbf{0.35}/0.11 & 2.42/0.92 & 1.02/\textbf{0.09} & \underline{0.59}/\textbf{0.09} \\
\texttt{skyway}     & 0.77/\underline{0.08} & \textbf{0.30}/0.09 & 1.55/0.60 & 1.24/\textbf{0.05} & \underline{0.73}/\textbf{0.05} \\
\texttt{freeway}    & \underline{0.46}/\underline{0.12} & \textbf{0.24}/\textbf{0.09} & 1.24/0.38 & 0.75/0.13 & 0.60/0.13 \\
\midrule
\textbf{average}    & 0.65/\underline{0.11} & \textbf{0.30}/\textbf{0.10} & 1.65/0.66 & 0.78/0.12 & \underline{0.56}/0.12 \\
\bottomrule
\multicolumn{6}{l}{\scriptsize KITTI odometry metric reported as \textit{XX / YY} with \textit{XX} [\%] and \textit{YY} [$\si{\degree}/100\,\si{\m}$] the translation and orientation errors, respectively.}

\end{tabularx}
\end{table*}

\begin{table}
    \centering
    \caption{Per-route vertical body velocity errors [m/s] with and without the proposed approximation (\textbf{best}).}
\label{tab:vel}
    \setlength{\tabcolsep}{2pt}
    \begin{tabularx}{\linewidth}{l|YY|YY}
\toprule
\multirow{2}{*}{\textbf{Seq. type}} & \multicolumn{2}{c|}{\textbf{3DRO} No $\velz{}{}$} & \multicolumn{2}{c}{\textbf{3DRO} (ours)}\\
 & RMSE & Avg. err. & RMSE & Avg. err \\
%\multirow{2}{*}{\textbf{Seq. type}} & \textbf{LTR/ gyro} & \textbf{2Fast-} & \textbf{3DRO} & \textbf{3DRO}\\
% & [3] & \textbf{2lamaa} [4] & No $\velz{}{}$ & (ours)\\
\midrule

\texttt{suburbs}    & 0.063 & 0.041 & \textbf{0.042} & \textbf{-0.003} \\
\texttt{industrial} & 0.056 & 0.029 & \textbf{0.044} & \textbf{-0.007} \\
\texttt{urban}      & 0.041 & 0.019 & \textbf{0.030} & \textbf{-0.002} \\
\texttt{forest}     & 0.094 & 0.069 & \textbf{0.063} & \textbf{-0.016} \\
\texttt{farm}       & 0.083 & 0.054 & \textbf{0.060} & \textbf{-0.013} \\
\texttt{regional}   & 0.082 & 0.055 & \textbf{0.047} & \textbf{-0.009} \\
\texttt{tunnel}     & 0.128 & 0.089 & \textbf{0.072} & \textbf{ 0.034} \\
\texttt{skyway}     & 0.245 & 0.209 & \textbf{0.129} & \textbf{ 0.100} \\
\texttt{freeway}    & 0.116 & 0.092 & \textbf{0.041} & \textbf{-0.009} \\
\bottomrule

\end{tabularx}
\end{table}

For SE(3) benchmarking, we include a version of LTR~\cite{are_we_ready_for} that leverages the gyroscope measurements in its cost function and 2Fast-2Lamaa~\cite{legentil20262fast} as baselines.
Both are lidar-inertial methods.
Fig.~\ref{fig:odom_traj} illustrates results obtained on the \texttt{2025-01-08-12-28} sequence (\texttt{suburbs}).
Table~\ref{tab:odomse3} reports the per-route average performance of the three methods, and a full per-sequence breakdown is provided in Appendix~\ref{app:se3details}.
Overall, 3DRO is outperformed by the two lidar-based methods on most sequence types, but the difference between LTR and 3DRO is relatively small, and even in favour of 3DRO on feature-deprived environments such as the \texttt{tunnel} and \texttt{skyway}.
Interestingly, 3DRO provides the best results out of the three methods on the \texttt{forest} sequences.
These sequences are particularly challenging for lidar-based methods due to the presence of dense unstructured vegetation.
We believe that the Doppler-based velocity constraint of DRO is less affected by the nature of the environment, and thus enables 3DRO to perform better in such conditions.

\begin{figure*}
    \centering
    \includegraphics[width=0.99\linewidth]{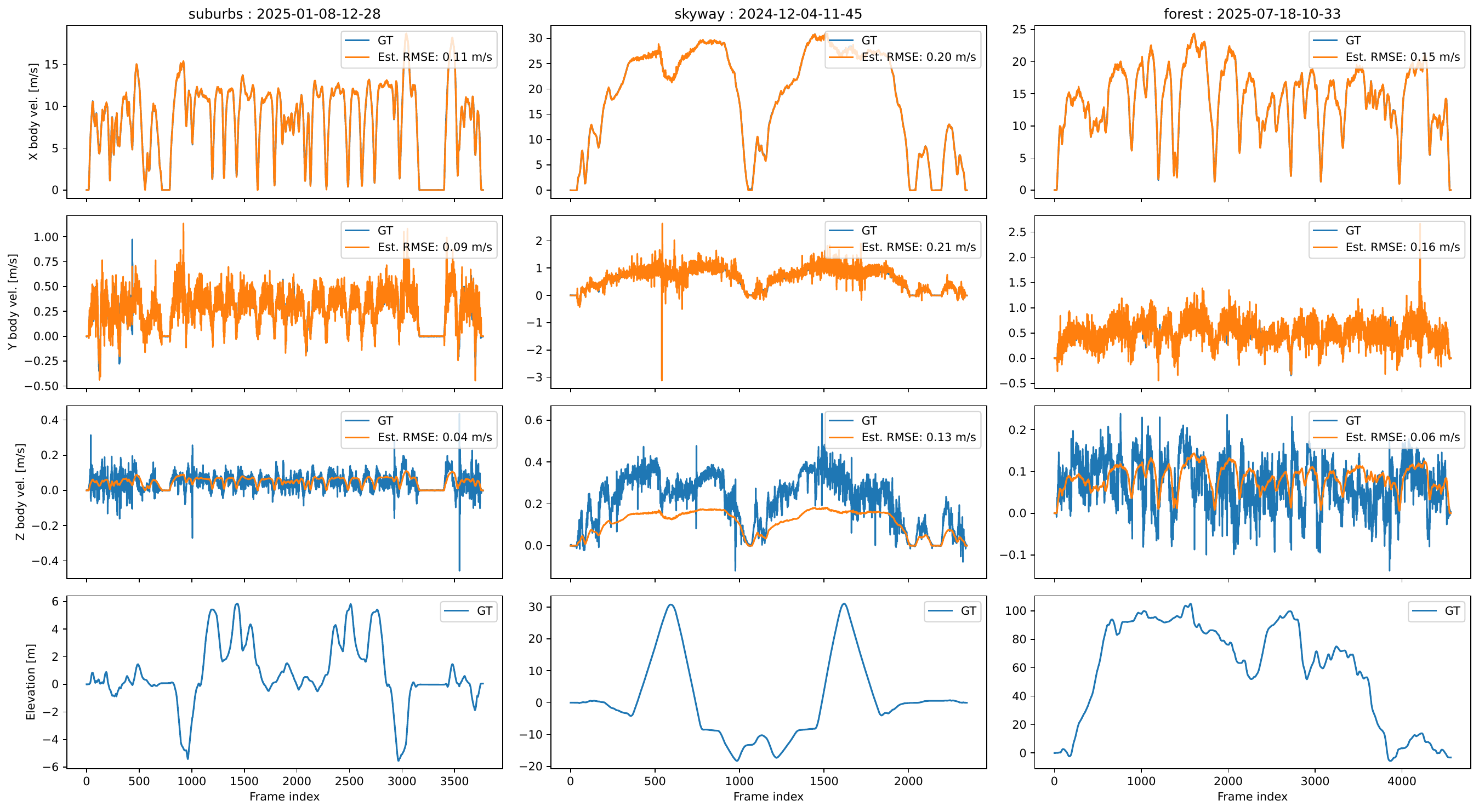}
    \caption{Per-axis velocity estimates obtained with 3DRO on three different sequences of the Boreas-RT dataset.}
    \label{fig:res_velocities}
\end{figure*}

While many driving environments can be considered `flat', SE(2) odometry leads to poor SE(3) localization estimates.
To show the added value of leveraging the gyroscope's three axes for integration and going beyond the SE(2) assumption, we also compute the SE(3) odometry errors using DRO's SE(2) output~\cite{legentil2025dro}, which only uses yaw angular velocities.
The SE(3) errors of DRO are significantly higher than those of 3DRO; between three and nine times higher, depending on the route.
This confirms that even `flat' environments such as the \texttt{suburbs} route can greatly benefit from SE(3) integration.
It is interesting to note that the highest error occurs on the \texttt{forest} sequences, where 3DRO performs the best.
These sequences contain the highest elevation changes, as illustrated in the bottom row of Fig.~\ref{fig:res_velocities}.

As an ablation study, we provide 3DRO's SE(3) odometry error when ignoring the vertical velocity of the sensor (3DRO No $\velz{}{}$) in Table~\ref{tab:odomse3}.
To further demonstrate the soundness of the proposed calibration of the sensor's vertical velocity with a simple constant multiplied by the estimated 2D velocity norm, we display the velocities obtained for three sequences in Fig.~\ref{fig:res_velocities}, and report the per-route average vertical body velocity error and \ac{rmse} in Table~\ref{tab:vel}.
While the velocity estimates are not perfect, as shown with a relatively high \ac{rmse}, the mean error is greatly improved.
This translates into greatly improved odometry results, reducing the error by up to \SI{40}{\percent}, empirically showing the effectiveness of our simple approximation.
We believe that the lower accuracy of the vertical velocity approximation for the \texttt{skyway} sequence in Fig.~\ref{fig:res_velocities} is due to a different payload of the vehicle: most sequences, including the one used for calibration, were collected with one or two people in the front seats, while four people were present in the car for \texttt{skyway} and and part of the \texttt{tunnel} sequences.
This difference impacts the average pitch of the vehicle's body, thus the validity of our approximation.

For reference, Fast-LIO2~\cite{xu2022fastlio2} and MOLA~\cite{blanco2025mola} results reported in~\cite{legentil20262fast} for a subset of the Boreas-RT dataset lead to per-route average errors from \SI{0.74}{\percent} to \SI{1.72}{\percent} on the \texttt{suburbs}, \texttt{regional}, and \texttt{tunnel} sequences.
3DRO clearly outperforms these lidar-based methods with results between \SI{0.31}{\percent} and \SI{0.59}{\percent} on the same sequences.

\subsection{SE(2) odometry results}

For SE(2) benchmarking, the Boreas-RT evaluation script projects the SE(3) pose estimates to SE(2) by ignoring the vertical component of the translation and the roll and pitch components of the rotation.
It essentially corresponds to the projection of the real 6-\ac{dof} motion to an `evaluation plane'.
In mostly planar environments, we expect the SE(2) odometry performance of 3DRO to be similar to that of DRO due to the small vertical component of the motion.
However, for routes with significant elevation changes, such as \texttt{forest} (see the elevation profile in Fig.~\ref{fig:res_velocities}), the SE(2) odometry estimate tends to overestimate the SE(2) translation as the true sensor velocity is not contained in the evaluation plane. 
In such scenarios, we anticipate a notable performance gap between 3DRO and DRO.
Table~\ref{tab:odomse2} reports the SE(2) odometry performance of 3DRO and compares it to RTR~\cite{are_we_ready_for} with additional angular velocity residuals, OG~\cite{legentil2025dowe}, and DRO~\cite{legentil2025dro}.
Both RTR and DRO are radar-based methods, while OG relies on wheel odometry and gyroscope integration.
RTR is the only method that does not directly integrate the gyroscope measurements for its rotational state estimation.

Overall, OG shows the best performance on most sequence types.
As expected, 3DRO provides slightly better estimates than DRO for all the sequences except the \texttt{forest} ones, where the gap is significantly larger.
This confirms our hypothesis that SE(3) odometry improves the SE(2) performance by accounting for non-planar motion.
It is important to note that this performance gain over DRO requires very little additional computation. The SE(3) integration~\eqref{eq:se3int} is performed in less than \SI{1}{\ms} per scan (on a single CPU core), while DRO's velocity estimates require around \SI{90}{\ms} on an NVIDIA RTX 5000 Mobile GPU.
This enables real-time operations as the radar scanning frequency is \SI{4}{\hertz}.

\begin{table}
    \centering
    \caption{Per-route average SE(2) odometry results on the Boreas-RT dataset (\textbf{best} and \underline{second best}).}
    \setlength{\tabcolsep}{2pt}
    \begin{tabularx}{\linewidth}{lYYYY}
\label{tab:odomse2}\\
\toprule
\multirow{2}{*}{\textbf{Sequence type}} & \textbf{RTR w/} & \textbf{OG} & \textbf{DRO} & \textbf{3DRO} \\
 & \textbf{gyro} \cite{are_we_ready_for} & \cite{legentil2025dowe} & \cite{legentil2025dro} & (ours) \\
\midrule

% ---------------------- SUBURBS ----------------------
\texttt{suburbs} & 0.36/0.05 & 0.25/\underline{0.04} & \underline{0.20}/\underline{0.04} & \textbf{0.18/0.03} \\
% ---------------------- INDUSTRIAL ----------------------
\texttt{industrial} & 0.45/0.08 & \textbf{0.28}/\underline{0.05} & 0.34/\underline{0.05} & \underline{0.33}/\textbf{0.03} \\
% ---------------------- URBAN ----------------------
\texttt{urban} & 1.10/0.16 & \textbf{0.44}/\underline{0.11} & 0.65/\underline{0.11} & \underline{0.63}/\textbf{0.10} \\
% ---------------------- FOREST ----------------------
\texttt{forest} & 0.57/\underline{0.06} & 0.38/\underline{0.06} & \underline{0.28}/\underline{0.06} & \textbf{0.18/0.02} \\
% ---------------------- FARM ----------------------
\texttt{farm} & 1.04/\underline{0.05} & \textbf{0.30}/\underline{0.05} & 0.40/\underline{0.05} & \underline{0.38}/\textbf{0.04} \\
% ---------------------- REGIONAL ----------------------
\texttt{regional} & 0.43/\underline{0.04} & \textbf{0.20}/\underline{0.04} & 0.26/\underline{0.04} & \underline{0.24}/\textbf{0.03} \\
% ---------------------- TUNNEL ----------------------
\texttt{tunnel} & 1.28/0.05 & \textbf{0.22}/\underline{0.04} & 0.35/\underline{0.04} & \underline{0.34}/\textbf{0.03} \\
% ---------------------- SKYWAY ----------------------
\texttt{skyway} & 0.90/\underline{0.03} & \textbf{0.14}/\textbf{0.02} & 0.41/\textbf{0.02} & \underline{0.40}/\textbf{0.02} \\
% ---------------------- FREEWAY ----------------------
\texttt{freeway} & 0.95/\underline{0.06} & \textbf{0.37}/\textbf{0.04} & 0.46/\textbf{0.04} & \underline{0.45}/\textbf{0.04}\\
\midrule
\textbf{average} & 0.79/0.06 & \textbf{0.29}/\underline{0.05} & 0.37/\underline{0.05} & \underline{0.35}/\textbf{0.04}
\\
\bottomrule
\multicolumn{5}{l}{\scriptsize KITTI odometry metric reported as \textit{XX / YY} with \textit{XX} [\%] and \textit{YY} [$\si{\degree}/100\,\si{\m}$]}
\vspace{-0.1cm}
\\
\multicolumn{5}{l}{\scriptsize the translation and orientation errors, respectively.}

\end{tabularx}
\end{table}

\section{Conclusion}

In this paper, we introduced 3DRO, a straightforward yet effective method to perform SE(3) odometry using both a 2D imaging radar and a 3-\ac{dof} gyroscope.
It relies on 2D-based direct radar odometry for in-plane velocity estimation.
After a simple calibration procedure, the out-of-plane velocity component is approximated using solely the previously mentioned 2D velocities.
SE(3) estimation is performed as the integration of the resulting 3D velocities and 3D gyroscope measurements.

This work demonstrates that ground robots do not require a 3D exteroceptive sensor to enable SE(3) state estimation.
Our experimental setup, using close to 650km of data from the Boreas-RT dataset, shows that 3DRO’s accuracy can compete with state-of-the-art lidar-based frameworks.
Future work includes the derivation of a more complex model to estimate the $z$ component of the radar velocity using the variation of DRO's velocity estimates and the pitch and roll angular velocities.

\bibliographystyle{IEEEtran}
\bibliography{references}

\appendix

\subsection{Boreas-RT SE(3)results details}

\label{app:se3details}

This appendix provides a full breakdown of the SE(3) odometry results on the Boreas-RT dataset in Table~\ref{tab:boreasrt_odom_structured} for the \emph{Structured} and \emph{Rural} sequences, and in Table~\ref{tab:boreasrt_odom_freeway} for the \emph{Highway} sequences.

\begin{table*}
    \centering
    \caption{Per-sequence SE(3)odometry results on the \emph{Structured} and \emph{Rural} sequences of the Boreas-RT dataset.}
    \setlength{\tabcolsep}{2pt}
    \begin{tabularx}{\linewidth}{lYYYY}
\label{tab:boreasrt_odom_structured}\\
\toprule
\textbf{Sequence type} & \textbf{ID} & \textbf{LTR w/ gyro} \cite{are_we_ready_for} & \textbf{2Fast-2Lamaa} \cite{legentil20262fast} & \textbf{3DRO} (ours) \\
\midrule

% ---------------------- SUBURBS ----------------------
\texttt{suburbs} & 2024-12-03-12-54 & 0.30/0.10 & 0.24/0.08 & 0.35/0.11 \\
& 2024-12-05-14-25 & 0.29/0.09 & 0.21/0.08 & 0.34/0.10 \\
& 2025-01-08-10-59 & 0.30/0.09 & 0.23/0.08 & 0.32/0.10 \\
& 2025-01-08-11-22 & 0.28/0.09 & 0.21/0.07 & 0.26/0.07 \\
& 2025-01-08-12-28 & 0.26/0.08 & 0.21/0.08 & 0.37/0.08 \\
& 2025-02-15-16-58 & 0.27/0.09 & 0.20/0.07 & 0.29/0.09 \\
& 2025-02-15-17-19 & 0.27/0.09 & 0.22/0.08 & 0.30/0.08 \\
& 2025-02-21-14-51 & 0.34/0.10 & 0.24/0.08 & 0.35/0.08 \\
& 2025-02-22-11-32 & 0.29/0.09 & 0.25/0.09 & 0.25/0.08 \\
& 2025-02-22-12-26 & 0.30/0.10 & 0.22/0.08 & 0.32/0.13 \\
\cmidrule{2-5}
& \textbf{Average} & 0.29/0.09 & \textbf{0.22/0.08} & 0.31/0.09 \\
\midrule
% ---------------------- INDUSTRIAL ----------------------
\texttt{industrial} 
& 2024-12-05-14-12 & 0.23/0.08 & 0.21/0.08 & 0.37/0.10 \\
& 2024-12-23-16-27 & 0.27/0.10 & 0.21/0.08 & 0.65/0.16 \\
& 2024-12-23-16-44 & 0.24/0.09 & 0.20/0.08 & 0.65/0.17 \\
& 2024-12-23-17-01 & 0.26/0.09 & 0.19/0.07 & 0.31/0.10 \\
& 2024-12-23-17-18 & 0.24/0.09 & 0.19/0.07 & 0.45/0.13 \\
\cmidrule{2-5}
& \textbf{Average} & 0.25/0.09 & \textbf{0.20/0.07} & 0.49/0.13 \\
\midrule
% ---------------------- URBAN ----------------------
\texttt{urban} 
& 2025-08-06-06-33 & 0.41/0.16 & 0.35/0.11 & 0.49/0.20 \\
& 2025-08-06-07-05 & 0.46/0.17 & 0.40/0.11 & 0.67/0.24 \\
& 2025-08-06-07-41 & 0.43/0.17 & 0.37/0.11 & 0.78/0.26 \\
& 2025-08-06-08-35 & 0.55/0.22 & 0.53/0.11 & 1.02/0.39 \\
& 2025-08-06-10-48 & 0.48/0.21 & 0.31/0.10 & 0.75/0.31 \\
& 2025-08-06-11-32 & 0.57/0.23 & 0.39/0.10 & 1.27/0.41 \\
& 2025-08-06-12-20 & 0.51/0.24 & 0.29/0.09 & 1.45/0.45 \\
\cmidrule{2-5}
& \textbf{Average} & 0.49/0.20 & \textbf{0.38/0.10} & 0.92/0.32 \\
\midrule

% ---------------------- FOREST ----------------------
\texttt{forest} 
& 2025-07-18-10-33 & 0.66/0.14 & 0.46/0.18 & 0.31/0.08 \\
& 2025-07-18-11-00 & 0.71/0.10 & 0.46/0.18 & 0.33/0.08 \\
& 2025-07-18-11-25 & 0.64/0.10 & 0.48/0.18 & 0.37/0.08 \\
& 2025-07-18-11-53 & 0.78/0.10 & 0.47/0.19 & 0.36/0.09 \\
\cmidrule{2-5}
& \textbf{Average} & 0.70/0.11 & 0.47/0.18 & \textbf{0.34/0.08} \\
\midrule

% ---------------------- FARM ----------------------
\texttt{farm} 
& 2025-07-18-14-55 & 0.45/0.11 & 0.35/0.10 & 0.47/0.12 \\
& 2025-07-18-15-12 & 0.44/0.10 & 0.37/0.10 & 0.58/0.11 \\
& 2025-07-18-15-30 & 0.42/0.11 & 0.38/0.10 & 0.50/0.14 \\
& 2025-07-18-15-48 & 0.60/0.11 & 0.39/0.11 & 0.59/0.13 \\
& 2025-07-18-16-05 & 0.49/0.10 & 0.40/0.11 & 0.43/0.10 \\
& 2025-08-13-09-01 & 0.40/0.13 & 0.33/0.10 & 0.54/0.16 \\
& 2025-08-13-09-21 & 0.40/0.13 & 0.34/0.09 & 0.62/0.23 \\
& 2025-08-13-09-46 & 0.42/0.13 & 0.34/0.09 & 0.57/0.17 \\
& 2025-08-13-10-12 & 0.42/0.13 & 0.35/0.10 & 0.56/0.18 \\
& 2025-08-13-10-36 & 0.40/0.12 & 0.34/0.10 & 0.54/0.15 \\
\cmidrule{2-5}
& \textbf{Average} & 0.44/0.12 & \textbf{0.36/0.10} & 0.54/0.15 \\
\bottomrule
\multicolumn{5}{l}{\scriptsize KITTI odometry metric reported as \textit{XX / YY} with \textit{XX} [\%] and \textit{YY} [$\si{\degree}/100\,\si{\m}$]}
\vspace{-0.1cm}
\\
\multicolumn{5}{l}{\scriptsize the translation and orientation errors, respectively.}
\end{tabularx}
\end{table*}

\begin{table*}
    \centering
    \caption{Per-sequence SE(3) odometry results on the \emph{Highway} sequences of the Boreas-RT dataset.}
    \setlength{\tabcolsep}{2pt}
    \begin{tabularx}{\linewidth}{lYYYY}
\label{tab:boreasrt_odom_freeway}\\
\toprule
\textbf{Sequence type} & \textbf{ID} & \textbf{LTR w/ gyro} \cite{are_we_ready_for} & \textbf{2Fast-2Lamaa} \cite{legentil20262fast} & \textbf{3DRO} (ours) \\
\midrule
% ---------------------- REGIONAL ----------------------
\texttt{regional} 
& 2024-12-03-13-13 & 0.34/0.09 & 0.18/0.07 & 0.47/0.09 \\
& 2024-12-03-13-34 & 0.34/0.09 & 0.20/0.08 & 0.36/0.08 \\
& 2024-12-10-12-07 & 0.34/0.09 & 0.20/0.08 & 0.34/0.07 \\
& 2024-12-10-12-24 & 0.35/0.10 & 0.21/0.08 & 0.30/0.08 \\
& 2024-12-10-12-38 & 0.32/0.08 & 0.18/0.08 & 0.34/0.07 \\
& 2024-12-10-12-56 & 0.38/0.10 & 0.21/0.08 & 0.52/0.10 \\
\cmidrule{2-5}
& \textbf{Average} & 0.35/0.09 & \textbf{0.20/0.08} & 0.39/\textbf{0.08} \\
\midrule

% ---------------------- TUNNEL ----------------------
\texttt{tunnel}
& 2024-12-04-14-28 & 2.26/0.12 & 0.27/0.08 & 0.63/0.17 \\
& 2024-12-04-14-34 & X         & 0.60/0.17 & 0.84/0.07 \\
& 2024-12-04-14-38 & 2.82/0.11 & 0.29/0.10 & 0.59/0.12 \\
& 2024-12-04-14-44 & 1.69/0.08 & 0.40/0.14 & 0.88/0.04 \\
& 2024-12-04-14-50 & X         & 0.30/0.10 & 0.31/0.08 \\
& 2024-12-04-14-59 & 2.05/0.09 & 0.37/0.11 & 0.35/0.06 \\
& 2024-12-04-15-04 & 2.36/0.11 & 0.26/0.09 & 0.30/0.09 \\
& 2024-12-04-15-10 & 1.02/0.09 & 0.33/0.11 & 0.55/0.07 \\
& 2024-12-04-15-19 & 2.87/0.11 & 0.28/0.09 & 0.63/0.10 \\
& 2024-12-04-15-24 & 1.41/0.09 & 0.39/0.14 & 0.81/0.06 \\
\cmidrule{2-5}
& \textbf{Average} & 2.06/0.10 & \textbf{0.35}/0.11 & 0.59/\textbf{0.09} \\
\midrule

% ---------------------- SKYWAY ----------------------
\texttt{skyway} 
& 2024-12-04-11-45 & 1.62/0.08 & 0.30/0.10 & 0.63/0.05 \\
& 2024-12-04-11-56 & 0.47/0.07 & 0.32/0.10 & 0.92/0.05 \\
& 2024-12-04-12-08 & 0.63/0.08 & 0.26/0.09 & 0.64/0.04 \\
& 2024-12-04-12-19 & 0.51/0.08 & 0.34/0.10 & 0.76/0.05 \\
& 2024-12-04-12-34 & 0.60/0.08 & 0.27/0.09 & 0.69/0.04 \\
\cmidrule{2-5}
& \textbf{Average} & 0.77/0.08 & \textbf{0.30}/0.09 & 0.73/\textbf{0.05} \\
\midrule

% ---------------------- FREEWAY ----------------------
\texttt{freeway} 
& 2025-07-18-16-24 & 0.53/0.13 & 0.26/0.10 & 0.65/0.17 \\
& 2025-08-13-07-54 & 0.42/0.11 & 0.20/0.07 & 0.46/0.09 \\
& 2025-08-13-11-52 & 0.44/0.13 & 0.26/0.10 & 0.70/0.14 \\
\cmidrule{2-5}
& \textbf{Average} & 0.46/0.12 & \textbf{0.24/0.09} & 0.60/0.13 \\
\bottomrule
\multicolumn{5}{l}{\scriptsize KITTI odometry metric reported as \textit{XX / YY} with \textit{XX} [\%] and \textit{YY} [$\si{\degree}/100\,\si{\m}$]}
\vspace{-0.1cm}
\\
\multicolumn{5}{l}{\scriptsize the translation and orientation errors, respectively.}

\end{tabularx}
\end{table*}

\end{document}